\documentclass[12pt]{article}

\usepackage{times,epsfig,graphics,amssymb,latexsym,amsmath,setspace}
\usepackage{array,threeparttable,hyperref,url}
\usepackage[usenames]{color}
\usepackage[authoryear,sort]{natbib}
\usepackage{paralist}
\usepackage{booktabs} 
\usepackage{multirow}
\usepackage{subfig} 
\usepackage{graphicx} 
\usepackage{parskip} 
\usepackage{hyperref}
\allowdisplaybreaks
\usepackage{booktabs}
\usepackage{siunitx}
\usepackage{hyperref}
\usepackage{algorithm,algorithmic}

\def\wt{\widetilde}

\def\boxit#1{\vbox{\hrule\hbox{\vrule\kern6pt
          \vbox{\kern6pt#1\kern6pt}\kern6pt\vrule}\hrule}}

\def\bse{\begin{eqnarray*}}
\def\ese{\end{eqnarray*}}
\def\be{\begin{eqnarray}}
\def\ee{\end{eqnarray}}
\def\bq{\begin{equation}}
\def\eq{\end{equation}}

\def\trans{^{\rm T}}

\newtheorem{proposition}{Proposition}

\newcommand{\blem}{\begin{lemma}}
\newcommand{\elem}{\end{lemma}}
\newcommand{\bthe}{\begin{theorem}}
\newcommand{\ethe}{\end{theorem}}

\newtheorem{lemma}{Lemma}
\newtheorem{theorem}{Theorem}

\def\bfd{{\bf d}}

\def\bfP{{\bf P}}

\def\bfS{{\bf S}}

\def\bfT{{\bf T}}

\def\blambda{{\mbox{\boldmath $\lambda$}}}
\def\bLambda{{\mbox{\boldmath $\Lambda$}}}

\def\delete#1{\iffalse #1 \fi}

\def\bse{\begin{eqnarray*}}
\def\ese{\end{eqnarray*}}
\def\bee{\begin{enumerate}}
\def\eee{\end{enumerate}}
\def\bqe{\begin{eqnarray}}
\def\eqe{\end{eqnarray}}
\def\bed{\begin{description}}
\def\eed{\end{description}}
\def\bei{\begin{itemize}}
\def\eei{\end{itemize}}

\def\argmin{\mathop{\rm argmin}}

\def\pmb#1{\setbox0=\hbox{#1}%
    \kern-.025em\copy0\kern-\wd0
    \kern.05em\copy0\kern-\wd0
    \kern-.025em\raise.0433em\box0 }
\def\pmbh#1#2{\setbox0=\hbox{#1}%
    \setbox1=\hbox{#2}%
    \kern-.025em\copy0\kern-\wd0
    \kern.05em\copy1\kern-\wd0
    \kern-.025em\raise.0433em\box0 }

\def\frac#1#2{{#1\over#2}}

\def\boxit#1{\vbox{\hrule\hbox{\vrule\kern6pt
   \vbox{\kern6pt#1\kern6pt}\kern6pt\vrule}\hrule}}

\def\listing#1{\vskip 4mm\begin{verbatim}\input#1 \vskip 4mm}
\def\thick#1{\hbox{\rlap{$#1$}\kern0.25pt\rlap{$#1$}\kern0.25pt$#1$}}

\def\wt{\widetilde}

\def\bfd{{\bf d}}

  \def\bfP{{\bf P}}

  \def\bfS{{\bf S}}
  \def\bfT{{\bf T}}

\def\bfzero{{\bf 0}}
\def\bfone{{\bf 1}}

\def\pmbh{{\pmb h}}

\def\calE{{\cal E}}

\def\calL{{\cal L}}

\def\mbfa{\mathbf{a}}  \def\mbfA{\mathbf{A}}
  \def\mbfB{\mathbf{B}}
  \def\mbfC{\mathbf{C}}
\def\mbfd{\mathbf{d}}  \def\mbfD{\mathbf{D}}
\def\mbfe{\mathbf{e}}  
  
\def\mbfg{\mathbf{g}}  \def\mbfG{\mathbf{G}}
  \def\mbfH{\mathbf{H}}
  \def\mbfI{\mathbf{I}}

  \def\mbfM{\mathbf{M}}
  \def\mbfN{\mathbf{N}}

  \def\mbfP{\mathbf{P}}
  \def\mbfQ{\mathbf{Q}}
  \def\mbfR{\mathbf{R}}
  \def\mbfS{\mathbf{S}}
  \def\mbfT{\mathbf{T}}
  
\def\mbfv{\mathbf{v}}  \def\mbfV{\mathbf{V}}
  
\def\mbfx{\mathbf{x}}  \def\mbfX{\mathbf{X}}
  
\def\mbfz{\mathbf{z}}  \def\mbfZ{\mathbf{Z}}

\renewcommand\today{\ifcase\month\or
   Jan\or Feb\or Mar\or Apr\or May\or
   Jun\or Jul\or Aug\or Sep\or Oct\or Nov\or
   Dec\fi
   \space\number\day, \number\year}

\def\boxit#1{\vbox{\hrule\hbox{\vrule\kern6pt\vbox{\kern6pt#1\kern6pt}\kern6pt\vrule}\hrule}}

\catcode`@=11
\def\@evenhead{\vbox{\hbox to \textwidth{\tiny \hfill \hfill \today } }}
\def\@oddhead{\vbox{\hbox to \textwidth{\tiny \hfill \hfill \today } }}

\def\argmin{\mathop{\rm argmin}}

\makeatletter
\g@addto@macro\normalsize{%
  \setlength{\abovedisplayskip}{6pt plus 2pt minus 2pt}
  \setlength{\belowdisplayskip}{6pt plus 2pt minus 2pt}
  \setlength{\abovedisplayshortskip}{0pt plus 2pt}
  \setlength{\belowdisplayshortskip}{6pt plus 2pt minus 2pt}
}
\makeatother

\begin{document}
\pagenumbering{arabic}
\setcounter{page}{1}
\baselineskip=14pt

\vskip 3mm

\date{}

  \title{\bf Sparse Convex Biclustering}
  \author{Jiakun Jiang, Dewei Xiang, Chenliang Gu\\
    School of Arts and Sciences, Beijing Normal University, Zhuhai, China\\
    Wei Liu\\
    School of Mathematics , Sichuan University, Chengdu, China\\
    Binhuan Wang\thanks{
    Corresponding author: wang.binhuan@gmail.com} \\
    Data \& Statistical Sciences, AbbVie Inc., Florham Park, NJ, USA \\}
  \maketitle

\begin{abstract}
Biclustering is an essential unsupervised machine learning technique for simultaneously clustering rows and columns of a data matrix, with widespread applications in genomics, transcriptomics, and other high-dimensional omics data. Despite its importance, existing biclustering methods struggle to meet the demands of modern large-scale datasets. The challenges stem from the accumulation of noise in high-dimensional features, the limitations of non-convex optimization formulations, and the computational complexity of identifying meaningful biclusters. These issues often result in reduced accuracy and stability as the size of the dataset increases. To overcome these challenges, we propose Sparse Convex Biclustering (SpaCoBi), a novel method that penalizes noise during the biclustering process to improve both accuracy and robustness. By adopting a convex optimization framework and introducing a stability-based tuning criterion, SpaCoBi achieves an optimal balance between cluster fidelity and sparsity. Comprehensive numerical studies, including simulations and an application to mouse olfactory bulb data, demonstrate that SpaCoBi significantly outperforms state-of-the-art methods in accuracy. These results highlight SpaCoBi as a robust and efficient solution for biclustering in high-dimensional and large-scale datasets.
\end{abstract}


\doublespacing

\section{Introduction}

In the rapidly evolving landscape of data-driven research, biclustering has emerged as a critical technique for analyzing complex data matrices by simultaneously clustering rows and columns. This dual partitioning capability distinguishes biclustering from traditional clustering approaches, often referred to as one-way clustering, which typically cluster observations based on all available features or cluster features across all observations. The ability to uncover submatrices where observations and features demonstrate synchronized patterns provides insights into context-specific relationships that would otherwise remain hidden in global analyses. This characteristic is particularly advantageous in a wide range of applications, especially within the domains of biological and biomedical data \citep{xie2019time}. These applications involve complex datasets derived from technologies such as single-cell RNA sequencing, which provide granular insights into cellular heterogeneity, disease-associated variant identification, and regulatory program inference. For instance, in gene expression studies, subsets of genes may exhibit co-expression only within specific cell types or experimental conditions, and biclustering can efficiently uncover these patterns, thereby offering more precise biological interpretations than traditional methods \citep{madeira2004biclustering, busygin2008biclustering}.

In our motivating example, we analyze data from the Mouse Olfactory Bulb (MOB) obtained through 10x Chromium single-cell RNA sequencing. This dataset consists of 305 observations, each of which contain 1,250 gene expressions. This case study is crucial, as understanding cell type heterogeneity and identifying marker genes are essential steps in elucidating the functional organization of this neural structure, which plays a fundamental role in processing olfactory information.

Despite its utility, traditional biclustering methods face significant challenges, particularly when applied to high-dimensional datasets typical of modern large-scale scientific inquiries. Earlier methodologies, often grounded in hybrid models or classical algorithms, made specific assumptions about data structures, which limited their flexibility and applicability to real-world data. Notable approaches based on singular value decomposition (SVD) \citep{lazzeroni2002plaid, bergmann2003iterative, turner2005improved} and those utilizing graph-based partitioning strategies have been noteworthy. However, their reliance on greedy optimization algorithms often leads to only local optima, thus limiting their efficacy in complex datasets \citep{chi2017convex, wang2023new}.

The advent of high-dimensional data exacerbates these challenges, as traditional non-convex optimization formulations struggle to meet the computational and analytical demands posed by modern datasets. Recent approaches have begun addressing these issues by incorporating sparsity into the biclustering process. Techniques based on sparse singular value decomposition (SVD) \citep{lee2010biclustering, sill2011robust, chen2013biclustering} attempt to improve results by enforcing penalties on singular values to achieve better feature selection. However, these methodologies often fall short in interpretability and cannot guarantee global optima due to the non-convex nature of their criterion functions \citep{helgeson2020biclustering}.

In response to these challenges, we propose Sparse Convex Biclustering (SpaCoBi) — an innovative framework that integrates sparsity into a convex optimization approach to effectively mitigate high-dimensional noise accumulation and facilitate precise feature selection. Motivied by the Sparse Convex Clustering algorithm \citep{wang2018sparse}, which simultaneously cluster observations and perform feature selection under a convex optimization framework with global optimum guaranteed, the proposed SpaCoBi algorithm incorporates sparsity-inducing lasso penalties within its biclustering model, which enhances the detection of true signals by suppressing irrelevant features. The computational strategy decomposes the problem into tractable subproblems, solved via a pseudo-regression scheme incorporating Sylvester-type updates, for which convergence is rigorously guaranteed. These novel contributions strategically address the inherent high-dimensional challenges, offering significant improvements in accuracy and robustness. Our approach offers several advantages: Firstly, in terms of accuracy and stability, the convex formulation of SpaCoBi allows for a unique global minimizer, significantly enhancing clustering precision across varying dimensions. Secondly, SpaCoBi's interpretability is improved by simultaneously estimating biclusters and selecting informative features, thus delineating biological insights such as cell subpopulations and their defining gene sets. Finally, computational efficiency is achieved through a straightforward iterative algorithm that leverages fast Sylvester solvers and warm starts, ensuring scalability.

The remainder of this paper is structured as follows: Section 2 details the SpaCoBi framework and its optimization algorithm. Section 3 discusses practical implementation considerations. Section 4 presents extensive simulations alongside a case study application on MOB data, illustrating the method's superior performance. Section 5 concludes with a summary and a discussion of future research directions. Technical details are deferred in Appendix.

\section{Sparse Convex Biclustering}

\subsection{Model}\label{model}

Let $\mbfX\in\mathbb{R}^{n\times p}$ be a data matrix with $n$ observations $X_{i\cdot}=(X_{i1},X_{i2},\dots,X_{ip})\trans$, with $p$ features, $i=1,\cdots,n$. We assume that the $n$ observations belong to $K$ unknown and non-overlapping classes, $C_1,\ldots, C_K$, and the $p$ features belong to $R$ unknown and non-overlapping classes, $D_1,\ldots, D_R$. To facilitate further derivations, we can also write the data matrix $\mbfX$ in feature-level as column vector $\mbfX=(\mbfx_{1},\cdots,\mbfx_{p})$, where $\mbfx_{j}=(X_{1j},\cdots,X_{nj})^{\trans}$, $j=1,\ldots,p$. Similarly we denote $\mbfA$ in feature-level as column vector $\mathbf{A}=(\mathbf{a}_{1},\cdots,\mathbf{a}_{p})$ and in observation-level as $(A_{1\cdot},\ldots, A_{n\cdot})\trans$. Define $\calE_1=\{l=(l_1,l_2):1\leq l_1<l_2 \leq n\}$ and $\calE_2=\{k=(k_1,k_2):0\leq k_1<k_2 \leq p \}$. Then denote $|\calE_1|$ and $|\calE_2|$ as the numbers of components of $\calE_1$ and $\calE_2$, respectively. We formulate the sparse convex biclustering problem as follows,
\begin{eqnarray}\label{obj_original0}
\min_{\mbfA \in \mathbb{R}^{n \times p}} && \frac{1}{2} \sum_{i=1}^n\| X_{i\cdot}- A_{i\cdot} \|_2^2  \\
 {\rm s.t.} && \sum_{l \in \calE_1} w_l \| A_{l_1\cdot} -A_{l_2\cdot} \|_q \leq t \nonumber\\
 &&  \sum_{k \in \calE_2} \wt{w}_k \| \mbfa_{k_1} -\mbfa_{k_2} \|_q \leq s \nonumber\\
 && \sum_{j=1}^p u_j \| \mbfa_j \|_2 \leq r,\nonumber
\end{eqnarray}
where the weights \( w_l \geq 0 \), \( \wt{w}_k \geq 0 \), and \( u_j \geq 0 \). The first and second constraints of \eqref{obj_original0} are designed to promote the integration of observations and features, respectively, for the purpose of biclustering. The third term emphasizes the importance of sparsity among the features. Here, the group LASSO or adaptive group LASSO penalty is deployed to select features, as it is common for the same feature to be shared by all observations.
By introducing additional slack variables \( \mbfv_l \), \( \mbfz_k \), and \( \mbfg_j \), which serve as essential components for applying the ADMM algorithm, we can reformulate the above problem into following equivalent constrained optimization problem,
\begin{eqnarray} \label{obj_original}
\min_{\mbfA \in \mathbb{R}^{n \times p}} && \frac{1}{2} \sum_{i=1}^n\| X_{i\cdot}- A_{i\cdot} \|_2^2 + \gamma_1 \sum_{l \in \calE_1} w_l \| \mbfv_l \|_q + \gamma_2 \sum_{k \in \calE_2} \wt{w}_k \| \mbfz_k \|_q + \gamma_3 \sum_{j=1}^p u_j \| \mbfg_j \|_2 \\
 {\rm s.t.} && A_{l_1\cdot} - A_{l_2\cdot} -\mbfv_l = \bfzero, \ \forall l \in \calE_1 \nonumber \\
 &&  \mbfa_{k_1} - \mbfa_{k_2} -\mbfz_k = \bfzero, \ \forall k \in \calE_2 \nonumber \\
 &&  \mbfa_j - \mbfg_j = \bfzero, \ j=1,\ldots, p. \nonumber
\end{eqnarray}
Then, the augmented Lagrangian problem is given by
\begin{eqnarray*}
&&\mathcal{L}_{\nu_1,\nu_2,\nu_3}(\mbfA,\mbfv,\mbfz,\mbfg, \bLambda_1,\bLambda_2,\bLambda_3) \\
&=&  \frac{1}{2} \sum_{i=1}^n\| X_{i\cdot}- A_{i\cdot} \|_2^2
+ \gamma_1 \sum_{l \in \calE_1} w_l \| \mbfv_l \|_q  + \gamma_2 \sum_{k \in \calE_2} \wt{w}_k \| \mbfz_k \|_q +\gamma_3 \sum_{j=1}^p u_j \| \mbfg_j \|_2\\
&& + \sum_{l \in \calE_1} \langle\blambda_{1l}, \mbfv_l -A_{l_1\cdot}+A_{l_2\cdot}\rangle+ \frac{\nu_1}{2} \sum_{l \in \calE_1} \| \mbfv_l -A_{l_1\cdot}+A_{l_2\cdot} \|_2^2 \\
&&  + \sum_{k \in \calE_2} \langle\blambda_{2k}, \mbfz_k -\mbfa_{k_1}+\mbfa_{k_2}\rangle + \frac{\nu_2}{2} \sum_{k \in \calE_2}\| \mbfz_k -\mbfa_{k_1}+\mbfa_{k_2} \|_2^2 \\
&& + \sum_{j=1}^p \langle\blambda_{3j}, \mbfg_j -\mbfa_j\rangle + \frac{\nu_3}{2} \sum_{j=1}^p \| \mbfg_j -\mbfa_j \|_2^2.
\end{eqnarray*}

\subsection{SpaCoBi Algorithm}

Minimizing the above augmented Lagrangian problem $\mathcal{L}_{\nu_1,\nu_2,\nu_3}(\mbfA,\mbfv,\mbfz,\mbfg, \bLambda_1,\bLambda_2,\bLambda_3)$ is challenging, but the ADMM algorithm enables us to iteratively update \( \mbfA, \mbfv, \mbfz, \mbfg, \bLambda_1, \bLambda_2, \) and \( \bLambda_3 \) in the following scheme:
\begin{eqnarray}
\mbfA^{m+1}  &=&  \argmin_{\mbfA}\calL_{\nu_1,\nu_2,\nu_3}(\mbfA,\mbfV^m,\mbfZ^m,\mbfG^m,\bLambda_1^m,\bLambda_2^m,\bLambda_3^m),\nonumber \\
\mbfV^{m+1}  &=&  \argmin_{\mbfV}\calL_{\nu_1,\nu_2,\nu_3}(\mbfA^{m+1},\mbfV,\mbfZ^m,\mbfG^m,\bLambda_1^m,\bLambda_2^m,\bLambda_3^m), \nonumber\\
\mbfZ^{m+1}  &=&  \argmin_{\mbfZ}\calL_{\nu_1,\nu_2,\nu_3}(\mbfA^{m+1},\mbfV^{m+1},\mbfZ,\mbfG^m,\bLambda_1^m,\bLambda_2^m,\bLambda_3^m), \nonumber \\
\mbfG^{m+1} &=&  \argmin_{\mbfG}\calL_{\nu_1,\nu_2,\nu_3}(\mbfA^{m+1},\mbfV^{m+1},\mbfZ^{m+1},\mbfG,\bLambda_1^m,\bLambda_2^m,\bLambda_3^m), \nonumber \\
\blambda_{1l}^{m+1}  &=&  \blambda_{1l}^{m}+\nu_1(\mbfv_{l}^{m+1}- A_{l_{1}\cdot}^{m+1}+A_{l_{2}\cdot}^{m+1}), \ l \in \calE_1,\nonumber \\
\blambda_{2k}^{m+1}  &=&  \blambda_{2k}^{m}+\nu_2(\mbfz_{k}^{m+1}- \mbfa_{k_{1}}^{m+1}+\mbfa_{k_{2}}^{m+1}), \ k \in \calE_2, \nonumber \\
\blambda_{3j}^{m+1}  &=&  \blambda_{3j}^{m}+\nu_3(\mbfg_{j}^{m+1}- \mbfa_{j}^{m+1}), \ j=1,\ldots,p. \nonumber
\end{eqnarray}

Next, we develop the detailed updating implementations for $\mbfA,\mbfV,\mbfZ,\mbfG, \bLambda_1, \bLambda_2, \bLambda_3$ in three steps. A summary of the SpaCoBi algorithm is shown in Algorithm \ref{algSpaCoBi}.

\textbf{Step 1 (update $\mbfA$)}: We need to minimize
\begin{eqnarray*}
f(\mbfA)&=&\frac{1}{2} \sum_{i=1}^n\| X_{i\cdot}- A_{i\cdot} \|_2^2 + \frac{\nu_1}{2} \sum_{l \in \calE_1} \| \wt{\mbfv}_l-A_{l_1 \cdot} + A_{l_2 \cdot} \|_2^2 \\
&& +\frac{\nu_2}{2} \sum_{k \in \calE_2} \| \wt{\mbfz}_k -\mbfa_{k_1}+\mbfa_{k_2} \|_2^2 +\frac{\nu_3}{2} \sum_{j=1}^p\| \wt{\mbfg}_j - \mbfa_j \|_2^2,
\end{eqnarray*}
where $\wt{\mbfv}_1=\mbfv_l + \frac{1}{\nu_1}\blambda_{1l}$, $\wt{\mbfz}_k=\mbfz_k + \frac{1}{\nu_2}\blambda_{2k}$, and $\wt{\mbfg}_j= \mbfg_j + \frac{1}{\nu_3} \blambda_{3j}$.

This step is the key component of the SpaCoBi algorithm. By applying matrix techniques, the estimate of $\mbfA$ can be obtained by solving the following equation with details deferred in Appendix:
\begin{eqnarray}
    \mbfM\mbfA + \mbfA\mbfN = \mbfH, \label{Stlvesterequation}
\end{eqnarray}
where
\begin{eqnarray*}
\mbfM &=&  \mbfI_n + \nu_1 \sum_{l \in \calE_1} (\mbfe_{l_1}-\mbfe_{l_2})(\mbfe_{l_1}-\mbfe_{l_2})\trans \\
\mbfN &=& \nu_2 \sum_{k \in \calE_2} (\mbfe_{k_1}^*-\mbfe_{k_2}^*) (\mbfe_{k_1}^*-\mbfe_{k_2}^*)\trans + \nu_3 \sum_{j=1}^p \mbfe_j^* (\mbfe_j^*)\trans \\
\mbfH&=& \mbfX + \sum_{l \in \calE_1} (\mbfe_{l_1}-\mbfe_{l_2})(\blambda_{1l}+ \nu_1\mbfv_l)\trans + \sum_{k \in \calE_2} (\blambda_{2k}+\nu_2\mbfz_k)(\mbfe_{k_1}^*-\mbfe_{k_2}^*)\trans + \sum_{j=1}^p (\blambda_{3j} + \nu_3 \mbfg_j) (\mbfe_j^*)\trans.
\end{eqnarray*}

If the edge sets $\calE_1$ and $\calE_2$ contain all possible edges, it is straightforward to verify
\begin{eqnarray*}
\sum_{l \in \calE_1} (\mbfe_{l_1}-\mbfe_{l_2})(\mbfe_{l_1}-\mbfe_{l_2})\trans &=& n\mbfI_n - \bfone_n \bfone_n\trans \\
\sum_{k \in \calE_2} (\mbfe_{k_1}^*-\mbfe_{k_2}^*) (\mbfe_{k_1}^*-\mbfe_{k_2}^*)\trans &=& p\mbfI_p - \bfone_p \bfone_p\trans.
\end{eqnarray*}
Then
\begin{eqnarray*}
\mbfM &=& (1+n\nu_1)\mbfI_n -\nu_1 \bfone_n \bfone_n\trans \\
\mbfN &=& p\nu_2 \mbfI_p - \nu_2 \bfone_p \bfone_p\trans + \nu_3 \mbfI_p.
\end{eqnarray*}
The equation (\ref{Stlvesterequation}) is a standard {\it Sylvester Equation}, which plays an important role
in control theory and many other branches of engineering. Its theoretical solution is based on eigenvector and eigenvalue decomposition \citep{Jameson1968} shown below, but it is computationally expensive. 

Assume $\mbfM$ has eigenvalues $\lambda_i, i=1,\ldots,n$, and $\mbfN$ has eigenvalues $\mu_j, j=1,\ldots,p$. Then, it is known that the equation (\ref{Stlvesterequation}) can be solved if and only if
$$\lambda_i+\mu_j \neq 0\ \ \ \ for\ \  all\ \  i,j.$$
 Here, \( \mbfM \) is positive definite and \( \mbfN \) is positive semi-definite, which implies that \( \mbfA \) is solvable. Assume that \( \mbfM \) and \( \mbfN \) can be diagonalized by orthogonal transformations:
\begin{eqnarray*}
\mbfT\trans\mbfM\mbfT=\left[\begin{array}{cccc}
\lambda_1 & & & \\
& \lambda_2 & & \\
& & \ddots & \\
& & & \lambda_n \\
\end{array}\right] \quad
\mbfS\trans\mbfN\mbfS=\left[\begin{array}{cccc}
\mu_1 & & & \\
& \mu_2 & & \\
& & \ddots & \\
& & & \mu_p \\
\end{array}\right].
\end{eqnarray*}
Then the solution is obtained as
$$\mbfA=\mbfS\wt{\mbfA}\mbfT\trans,$$
where $\wt{\mbfA}=(\wt{a}_{ij})$,
$$\wt{a}_{ij}=\frac{\wt{c}_{ij}}{\mu_i+\lambda_j},\ \wt{\mbfC}=(\wt{c}_{ij})=\bfS\trans\mbfC\bfT.$$

Alternatively, the Bartels-Stewart algorithm \citep{bartels1972solution} is the standard numerical solution that transforms the {\it Sylvester Equation} into a triangular system with the Schur decomposition and then solves it with forward or backward substitutions. In this manuscript, we implement a modified Bartels-Stewart algorithm proposed by \cite{Sorensen2003}, which is more computationally efficient.

\textbf{Step 2 (update $\mbfV$, $\mbfZ$ and $\mbfG$):} It is clear that the vectors $\wt{\mbfv}_l$, $\wt{\mbfz}_k$ and $\wt{\mbfg}_j$ are separable in the objective function, thus $\wt{\mbfv}_l$ and $\wt{\mbfz}_k$ can be solved by the proximal map:
\begin{eqnarray*}
\mbfv_l &=& \argmin_{\mbfv_l} \frac{1}{2} \|\mbfv_l - (A_{l_1\cdot}-A_{l_2\cdot} - \nu_1^{-1}\blambda_{1l}) \|_2^2
 + \frac{\gamma_1 w_l}{\nu_1} \| \mbfv_l\|_q \\
 &=& \textrm{prox}_{\sigma_{1l}\|\cdot\|_q} (A_{l_1\cdot}-A_{l_2\cdot}- \nu_1^{-1}\blambda_{1l}) \\
\mbfz_k &=& \argmin_{\mbfz_k} \frac{1}{2} \|\mbfz_k - (\mbfa_{k_1}-\mbfa_{k_2} - \nu_2^{-1}\blambda_{2k}) \|_2^2
 + \frac{\gamma_2 \wt{w}_k}{\nu_2} \| \mbfz_k\|_q \\
 &=& \textrm{prox}_{\sigma_{2k}\|\cdot\|_q} (\mbfa_{k_1}-\mbfa_{k_2} - \nu_2^{-1}\blambda_{2k}) \\
\mbfg_j &=& \argmin_{\mbfg_j} \frac{1}{2} \|\mbfg_j - (\mbfa_j - \nu_3^{-1}\blambda_{3j}) \|_2^2
 + \frac{\gamma_3 u_j}{\nu_3} \| \mbfg_j\|_2 \\
 &=& \textrm{prox}_{\sigma_{3j}\|\cdot\|_2} (\mbfa_j - \nu_3^{-1}\blambda_{3j})
\end{eqnarray*}
where $\sigma_{1l}=\gamma_1 w_l/\nu_1$, $\sigma_{2k}=\gamma_2 \wt{w}_k/\nu_2$ and $\sigma_{3j}=\gamma_3 u_j/\nu_3$. we refer the readers to table 1 in \cite{Chi2015} for the solutions to the proximal map of $L_q$-norm for $q=1,2$ and $\infty$ . In this article, the $L_2$-norm is primarily employed.

\textbf{Step 3 (update $\bLambda_1,\bLambda_2$ and $\bLambda_3$):} We update$\blambda_{1l}$, $\blambda_{2k}$ and $\blambda_{3j}$ by
\begin{eqnarray*}
\blambda_{1l} &\leftarrow& \blambda_{1l} + \nu_1(\mbfv_l - A_{l_1\cdot} + A_{l_2\cdot}), \\
\blambda_{2k} &\leftarrow& \blambda_{2k} + \nu_2(\mbfz_k - \mbfa_{k_1}+\mbfa_{k_2}), \\
\blambda_{3j} &\leftarrow& \blambda_{3j} + \nu_3(\mbfg_j - \mbfa_j).
\end{eqnarray*}

\begin{algorithm}[!htb]
\protect\caption{\quad{}SpaCoBi \label{algSpaCoBi}}

\begin{enumerate}
\item Initialize $\mbfV^{0}, \mbfZ^{0}, \mbfG^{0}, \bLambda_1^{0}, \bLambda_2^{0}$ and $\bLambda_3^{0}$. Calculate
\begin{eqnarray*}
\mbfM &=& (1+n\nu_1)\mbfI_n -\nu_1 \bfone_n \bfone_n\trans \\
\mbfN &=& p\nu_2 \mbfI_p - \nu_2 \bfone_p \bfone_p\trans + \nu_3 \mbfI_p.
\end{eqnarray*}
For $m=1,2,\ldots$
\item Solve the {\it Sylvester Equation} $\mbfM\mbfA + \mbfA\mbfN = \mbfH^{m-1}$ to obtain $\mbfA^m$, where
\begin{eqnarray*}
\mbfH^{m-1} &=& \mbfX + \sum_{l \in \calE_1} (\mbfe_{l_1}-\mbfe_{l_2})(\blambda_{1l}^{m-1}+ \nu_1\mbfv_l^{m-1})\trans + \\
&& \sum_{k \in \calE_2} (\blambda_{2k}^{m-1}+\nu_2\mbfz_k^{m-1})(\mbfe_{k_1}^*-\mbfe_{k_2}^*)\trans + \sum_{j=1}^p (\blambda_{3j}^{m-1} + \nu_3 \mbfg_j^{m-1}) (\mbfe_j^*)\trans.
\end{eqnarray*}
\item For $l\in\calE_1$, do
\begin{eqnarray*}
\mbfv_{l}^{m}= \textrm{prox}_{\sigma_{1l}\|\cdot\|_q} (A_{l_1\cdot}^{m}-A_{l_2\cdot}^{m}- \nu_1^{-1}\blambda_{1l}^{m-1}).
\end{eqnarray*}
\item For $k\in\calE_2$, do
\begin{eqnarray*}
\mbfz_{l}^{m}= \textrm{prox}_{\sigma_{2k}\|\cdot\|_q} (\mbfa_{k_1}^m-\mbfa_{k_2}^m - \nu_2^{-1}\blambda_{2k}^{m-1}).
\end{eqnarray*}
\item For $j=1,\ldots,p$, do
\begin{eqnarray*}
\mbfg_j^{m} = \textrm{prox}_{\sigma_{3j}\|\cdot\|_2} (\mbfa_j^{m} - \nu_3^{-1}\blambda_{3j}^{m}).
\end{eqnarray*}
\item For $l\in\calE_1$, $k \in\calE_2$ and $j=1,\ldots,p$, do
\begin{eqnarray*}
\blambda_{1l}^m &=& \blambda_{1l}^{m-1} + \nu_1(\mbfv_l^m - A_{l_1\cdot}^m + A_{l_2\cdot}^m)\\
\blambda_{2k}^m &=& \blambda_{2k}^{m-1} + \nu_2(\mbfz_k^m - \mbfa_{k_1}^m+\mbfa_{k_2}^m) \\
\blambda_{3j}^m &=& \blambda_{3j}^{m-1} + \nu_3(\mbfg_j^m - \mbfa_{j}^m).
\end{eqnarray*}

\item Repeat Steps 2-6 until convergence. \end{enumerate}
\end{algorithm}

\section{Implementation}
In this section, we discuss practical considerations for implementing the proposed algorithm, including algorithmic convergence and the selection of tuning parameters. 
\subsection{Algorithmic Convergence}

In the context of convex clustering, \cite{wang2023new} discussed the convergence of their convex biclustering algorithms. The primary difference between the objective function in (\ref{obj_original}) and that in \cite{wang2023new} is the additional group LASSO penalty term on the feature-level vectors. According to \cite{chi2017convex}, this remains a convex optimization problem, and under mild regularization conditions, the convergence of SpaCoBi algorithms is guaranteed.

\subsection{Selection of Weights}
{ In this section, we introduce the stragey of selecting { the weights \( w_{l} \), \(l\in \mathcal{E}_1\) and \( \wt{w}_{k} \), \(l\in \mathcal{E}_2\) for the fused-LASSO penalty, as well as the selection of the factor \( u_j \), \( j = 1, \ldots, p \), }in the group LASSO penalty. Following \cite{Chi2015}, we select the weights by combining the \( m \)-nearest neighbor method with the Gaussian kernel. Specifically, the weight \( w_l \) between samples \((l_1, l_2)\) is defined as:
\begin{equation}
w_{l} = t_{l_1, l_2}^{m} \exp\left(-\phi \lVert {X}_{l_1 \cdot} - {X}_{l_2\cdot} \rVert_2^2\right),
\end{equation}
where \( t_{l_1, l_2}^{m} \) is \( 1 \) if individual \( l_2 \) is within the \( m \)-nearest neighbor range of individual \( l_1 \), and \( 0 \) otherwise. Similarly, we can define $\wt{w}_k$. When $m$ is small, this choice of weights is applicable to a wide range of $\phi$. In our numerical results, $m$ is fixed at $5$, and $\phi$ is fixed at $0.5$. {The factor $u_j$ can be chosen as 
$
\frac{1}{\lVert \mathbf{a}_j^{(0)} \rVert_2}
$,
where $\mathbf{a}_j^{(0)}$ is the estimate of $\mathbf{a}_j$ in (\ref{obj_original}) when $\gamma_3=0$. }This factor selection method imposes a smaller penalty on informative features and a larger penalty on non-informative features, thereby enhancing the cluster accuracy and variable selection performance compared to its non-adaptive version. Finally, to ensure that the optimal tuning parameters \( \gamma_1 \), \( \gamma_2 \) and \( \gamma_3 \) remain within a relatively stable range, regardless of the dimension and sample size, the weights \( w_{l} \) ,  \( \wt{w}_{k} \) and factors \( u_j \) are rescaled to sum to \( \frac{1}{\sqrt{p}} \),\( \frac{1}{\sqrt{n}} \) and \( \frac{1}{\sqrt{n}} \), respectively. This rescaling facilitates the simultaneous consideration of both parameters during computation and does not affect the final clustering path.
}

\subsection{Selection of Tuning Parameters}
This section discusses the methods for selecting the tuning parameters \( \gamma_1 \), \( \gamma_2 \), and \( \gamma_3 \). Recall that \( \gamma_1 \) controls the number of observation-level clusters, \( \gamma_2 \) governs the number of feature-level clusters, and \( \gamma_3 \) regulates the sparsity of the feature vectors. Utilizing three separate tuning parameters to independently manage the number of row and column clusters, as well as sparsity, provides greater flexibility in the application of the algorithm.

We first demonstrate the effectiveness of the tuning parameter $\gamma_3$ in controlling the accuracy of variable selection through a numerical simulation. In this example, 60 samples with 400 features were generated from 4 classes, respectively. Among the 400 features, only 40 are informative for clustering, and the remaining 360 are noise. Refer to the detailed simulation setup in the section on numerical simulations. By fixing $\gamma_{1}$ and  $\gamma_{2}$ at a value of $50$ that is potentially close to the optimal one, and gradually increasing $\gamma_3$ from $e^0$ to $e^{7.5}$, we plotted the False Negative Rate (FNR) and False Positive Rate (FPR) paths of the final estimator. As shown in Figure (\ref{Fig.main1}), when $\gamma_3$ approaches zero, all features are included. When $\gamma_3$ increases to a certain range, all non-informative features are excluded, and all informative features are completely retained, meaning all useful variables are accurately selected. This demonstrates the sensitivity of $\gamma_3$ on the variable selection performance of the final estimator.

\begin{figure}
		\centering 
		\includegraphics[width=0.7\textwidth]{ 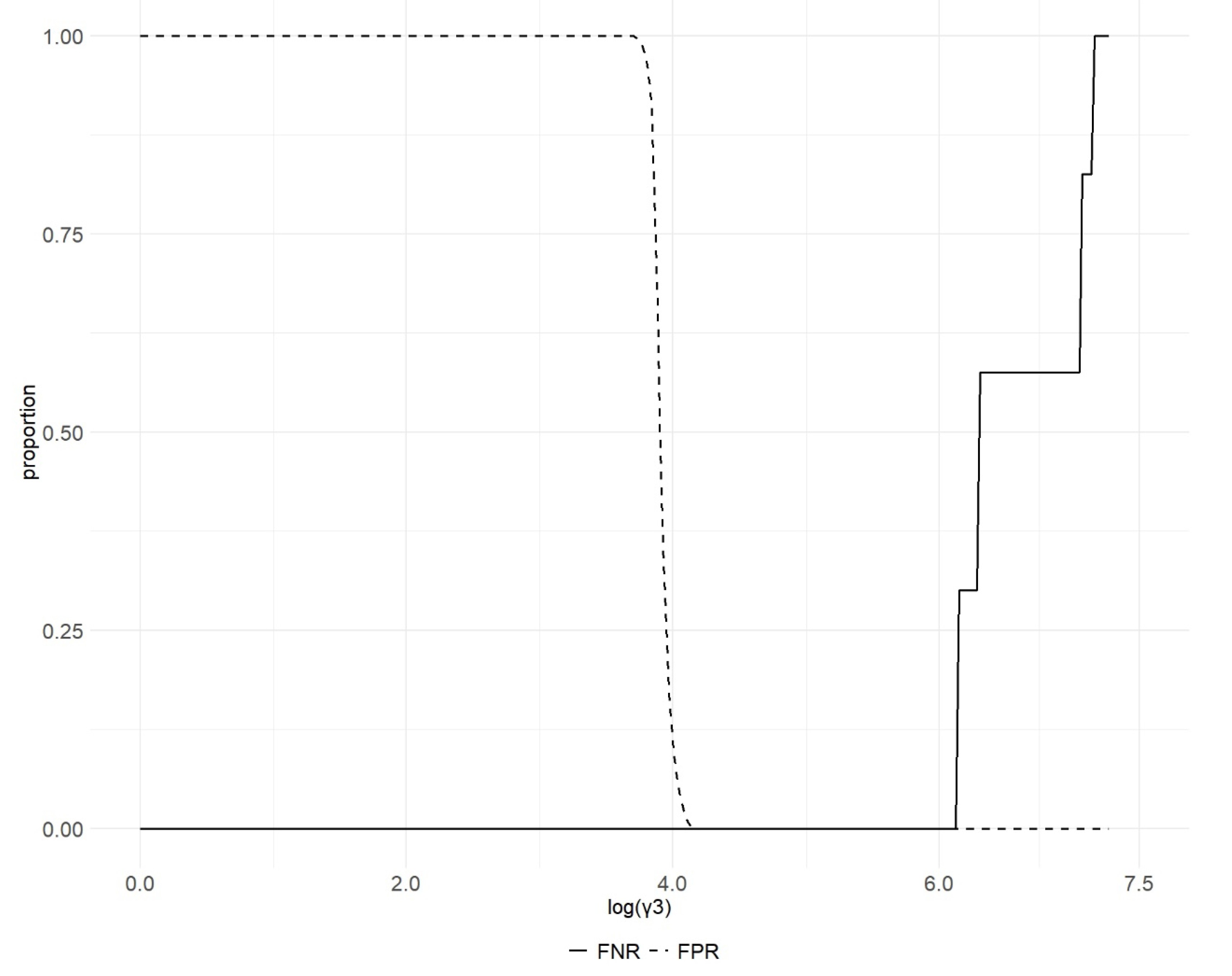} 
		\caption{Illustration of the effectiveness of $\gamma_3$  on variable selection accuracy. The solid curve is the path of false negative rate (FNR), and the dashed curve is the path of false positive rate (FPR).} 
		\label{Fig.main1} 
\end{figure}

Generally, if computational resources are sufficient, a three-dimensional grid search can provide an optimal set of tuning parameters under certain criteria. Considering the specificity of the tuning parameters $\gamma_1$ and $\gamma_2$, if the data matrix is large-scale,{the computational burden associated with three-dimensional grid search becomes prohibitively expensive}, we can adopt a similar strategy to that suggested by \cite{chi2017convex}: combining these two tuning parameters with appropriately rescaled penalty terms. This approach reduces the computational burden but necessitates clustering rows and columns in a proportional manner. Specifically, we rewrite the sparse convex biclustering problem as the following minimization problem with two tuning parameter:
\begin{eqnarray} \label{eq:tuning-para}
\min_{\mbfA \in \mathbb{R}^{p \times n}} && \frac{1}{2} \sum_{i=1}^n\| X_{i\cdot}- A_{i\cdot} \|_2^2 + \gamma \left\{ \sum_{l \in \calE_1} w_l \| \mbfv_l \|_q + \sum_{k \in \calE_2} \wt{w}_k \| \mbfz_k \|_q\right\} + \gamma_3 \sum_{j=1}^p u_j \| \mbfg_j \|_2 
\end{eqnarray}


Existing research has proposed various strategies for tuning parameter selection in biclustering. \cite{chi2017convex} proposed a hold-out validation method for convex biclustering by randomly selecting elements from the data matrix and using an estimated model based on the remaining elements to evaluate the quality of the predictions for the hold-out set. However, \cite{Fang2012}pointed out that data splitting reduces the size of the training dataset, making cross-validation methods inefficient. They proposed using stability selection in clustering analysis, which has also been adopted in subsequent clustering studies. \cite{wang2018sparse} and \cite{wang2023new} adopted stability selection in their sparse convex clustering and convex biclustering algorithms, respectively.  For sparse convex biclustering, we apply stability selection in a similar manner to tune $\gamma_1$, $\gamma_2$ and $\gamma_3$.

Specifically, for two bootstrap samples and a set of tuning parameters, the clustering algorithm can produce two biclustering results, each containing the centers and the number of clusters. Using these two biclustering results, a stability measure can be calculated to assess the consistency between the two clustering outcomes, which utilizes the Clustering Distance defined in \cite{Fang2012}:

{\bf Definition 1}: (Clustering distance) The distance between any two clustering $\psi_1(x)$ and $\psi_2(x)$ is defined as
$$
d_F\left(\psi_1, \psi_2\right)=E_{x^0 \sim F, y^0 \sim F}\left\{\left|I\left\{\psi_1\left(x^0\right)=\psi_1\left(y^0\right)\right\}-I\left\{\psi_2\left(x^0\right)=\psi_2\left(y^0\right)\right\}\right|\right\},
$$
where $I\{\cdot\}$ is the indicator function, and the expectation is taken over $x^0$ and $y^0$, two independent observations sampled from $F$.


\subsection{Warm-Start}
To alleviate the substantial computational burden associated with the SpaCoBi algorithm during multiple repeated simulations, we explored a ``Warm-Start" strategy. In machine learning and recommender systems, Cold-Start and Warm-Start refer to how systems handle challenges stemming from different stages of data availability. Cold-Start typically indicates a lack of historical data, while Warm-Start is a valuable optimization technique. Specifically, Warm-Start uses the optimal solution of a related or simplified problem as the initial value for the current, more complex problem. This high-quality initialization enables the optimizer to start closer to the global optimum, accelerating convergence and improving computational efficiency, particularly in non-linear optimization problems with multiple local minima.

In some biclustering literature, researchers employ aggressive computational schemes that use a good starting point and perform a single iteration to yield an approximate solution \citep{ramachandra2023optimization}. While this approach significantly increases calculation speed, we adopt a more methodical strategy. In this manuscript, we leverage the Warm-Start strategy by using the converged result of a previous optimization step in our grid search as the initial value for subsequent optimizations. This methodology yielded optimal computational results in our extensive numerical experiments.


We conducted an experiment that compared the computational time and the number of iterations for ten different grid search points at varying sample sizes \( n \), the total features \( p \), and the number of true informative features \( p_{\text{true}} \). For each sample, we performed \( 30 \) repetitions and calculated the average values. We fixed \( \gamma_1 \) and \( \gamma_2 \) at \( 50 \) (a potentially optimal parameter) while searching for the optimal \( \gamma_3 \) among \( 10 \) values ranging from \( 30 \) to \( 150 \). Both programs were executed with the same iterative convergence tolerance of \( e^{-5} \). Although tighter tolerances of \( e^{-6} \) to \( e^{-7} \) can yield marginally better results, a tolerance of \( e^{-5} \) is generally sufficient to obtain reasonably accurate clusters in most datasets. The times presented in the table are measured in seconds and represent the average duration for the programs to reach iterative convergence, with values in parentheses indicating the average number of iterations required for each convergence. The experiments were run on a computer equipped with an AMD Ryzen 5 4600H CPU and 16GB RAM.

\begin{table}[htbp]
\centering
\scriptsize
\caption{Computational Efficiency Analysis of SpaCoBi with Warm-Start (Time in Seconds and Average Iterations)}
\label{tab:warm_start_efficiency}
\begin{tabular}{c c c c c c c c c}
\toprule
\textbf{Parameters} & \multicolumn{3}{c}{\boldmath$n=60$} & \multicolumn{3}{c}{\boldmath$n=120$} & \multicolumn{2}{c}{\boldmath$n=200$} \\
\cmidrule(lr){2-4} \cmidrule(lr){5-7} \cmidrule(lr){8-9}
$p$ & 120 & 200 & 400 & 120 & 200 & 400 & 120 & 200 \\
$p_{\text{true}}$ & 40 & 40 & 40 & 40 & 40 & 40 & 40 & 40 \\
\midrule
\textbf{SpaCoBi} & 6.56 (112.8) & 10.56 (62.5) & 18.41 (32.3) & 13.59 (125.4) & 12.80 (40.5) & 27.04 (24.1) & 22.42 (121.4) & 13.96 (22.4) \\
\textbf{SpaCoBi (Warm-Start)} & 4.21 (57.3) & 8.71 (49.2) & 13.95 (23.9) & 8.05 (42.9) & 10.29 (33.3) & 17.81 (15.7) & 11.40 (33.6) & 10.16 (16.3) \\
\midrule
\textbf{Efficiency $\uparrow$} & {55.72\%} & {21.27\%} & {32.01\%} & {68.80\%} & {24.45\%} & {51.86\%} & {96.66\%} & {37.51\%} \\
\bottomrule
\end{tabular}
\end{table}

Upon comparing the time and number of iterations in Table \ref{tab:warm_start_efficiency}, we found that the warm start algorithm improved computational efficiency by at least \( 21.72\% \) compared to the SpaCoBi algorithm without the Warm-Start feature (For context, reducing iteration time from \( 100 \) seconds to \( 50 \) seconds represents a \( 100\% \) improvement in efficiency). These results clearly highlight the significant advantages of adopting the Warm-Start strategy. When handling real-world data requiring biclustering, practitioners often fix two parameters and search for repetitive penalty parameters, similar to our approach with the ten search points. In this common scenario, the Warm-Start method proves to be highly effective in accelerating the SpaCoBi program. Consequently, all subsequent numerical calculations presented in this manuscript utilize this enhanced method.

\section{Numerical Results}

This section is dedicated to demonstrating the superior performance of our proposed Sparse Convex Biclustering (SpaCoBi) method. The evaluation is conducted across two distinct domains: a comprehensive set of simulated examples (Subsection 4.1) and a real-world application involving Mouse Olfactory Bulb (MOB) gene expression data (Subsection 4.2). 

In simulation studies, each simulation was repeated \( 50 \) times to ensure robust statistical inference. The primary metric for quantifying the accuracy of the biclustering results is the Adjusted Rand Index (ARI), which measures the agreement between the estimated bicluster assignments and the true clustering labels, which are predefined and known in the simulation context. The ARI ranges from \(-1\) to \(1\), where a higher value indicates superior clustering performance. Given the true cluster labels, it is possible to evaluate the maximum potential performance of the candidate methods by tuning them to maximize the ARI. Additionally, the algorithm's capability to select features is rigorously assessed using the False Negative Rate (FNR) and the False Positive Rate (FPR). The numeric performance of the proposed SpaCoBi algorithm is compared to Bi-ADMM (\citep{wang2023new}) and COBRA \citep{chi2017convex} in terms of above metrics on biclustering problems.

To ensure a fair comparison with the Bi-ADMM (\(L_2\)) method, both algorithms were implemented using the formulation presented in Equation (\ref{eq:tuning-para}), which requires a two-dimensional grid search over the tuning parameters. For each repetition, the optimal set of tuning parameters is determined by maximizing the ARI on a validation data matrix that shares the same underlying classification structure as the training data but is distinct from it \citep{Witten.Tibshirani:2010}. 

\subsection{Simulation studies}

 We simulate a \( n \times p \) data matrix that consists of non-informative features and a checkerboard bicluster structure. This structure contains \( p_{\text{true}} \) informative features with non-zero means and \( p - p_{\text{true}} \) non-informative features.  For informative features, \( X_{ij} \) is generated as follows: we assign cluster indices to observations (rows) by randomly sampling the set \( \{1, \ldots, 4\} \), and cluster indices are assigned to features (columns) following a similar procedure. Consequently, for different runs, the generated data matrices and the classifications differ. For instance, with \( 60 \) observations divided into \( 4 \) classes, one run may produce \( 4 \) groups with \( 15 \) elements each, while the next run could yield class sizes of \( 10, 5, 5, \) and \( 20 \), respectively. The total number of biclusters is \( M = 4 \times 4 \), indicating that each \( X_{ij} \) belongs to one of these \( M \) biclusters. Then, random samples for each bicluster are generated from a normal distribution: \( X_{ij} \) i.i.d. \( \sim \mathcal{N}(\mu_{kr}, \sigma^2) \), where samples from row cluster \( k \in \{1, \ldots, 4\} \) and column cluster \( r \in \{1, \ldots, 4\} \) follow a normal distribution with mean \( \mu_{kr} \) and variance \( \sigma^2 \). The mean \( \mu_{kr} \) is chosen uniformly from the sequence \( \{-10, -9, \ldots, 9, 10\} \). Finally, the remaining \( p - p_{\text{true}} \) noise features are generated from \( \mathcal{N}(0, 9) \).

The results presented in Table \ref{tab:ari_comparison_reorganized} clearly demonstrate that the SpaCoBi method consistently outperforms the Bi-ADMM(\(L_2\)) algorithm across all tested sample size settings. Notably, as the dimensionality of features (\(p\)) increases, particularly in high-dimensional scenarios, the performance of the Bi-ADMM(\(L_2\)) algorithm—lacking a sparsity penalty—deteriorates rapidly. In contrast, the performance degradation of the SpaCoBi method is considerably less pronounced. Under large sample conditions, SpaCoBi exhibits significantly superior performance, underscoring the crucial role of an informative feature selection mechanism in the high-dimensional biclustering process. Furthermore, as long as non-informative features are present in the data, the Adjusted Rand Index (ARI) of SpaCoBi consistently surpasses that of the Bi-ADMM(\(L_2\)) algorithm.


From the Area Under the Curve (AUC) values presented in Table \ref{tab:fnr_fpr_auc_comparison_reorganized}, we observe that the SpaCoBi method can nearly perfectly identify the informative features, with the AUC approaching \( {0.8} \) under these simulation conditions. This high accuracy in feature selection is corroborated by low False Negative Rates (FNR) and low False Positive Rates (FPR). The reduced clustering accuracy in the Bi-ADMM(\(L_2\)) case can be directly attributed to the abundance of non-informative features, highlighting the necessity of selecting informative features and demonstrating the superior capability of the SpaCoBi algorithm's feature selection mechanism. This observation is consistent with similar findings reported by \cite{Tan2014} and \cite{wang2018sparse} in their studies on sparse convex clustering, which provided a critical theoretical foundation and motivation for the development of SpaCoBi.

\begin{table}[htbp]
\centering
\caption{Simulation results for SpaCoBi, Bi-ADMM, and COBRA in terms of the ARI, separated by Training (denoted as Train) and Validation (denoted as Val) sets.}
\label{tab:ari_comparison_reorganized}
\setlength{\tabcolsep}{3pt}
\begin{tabular}{
c 
c 
c 
S[table-format=1.2] 
S[table-format=1.2] 
S[table-format=1.2] 
S[table-format=1.2] 
S[table-format=1.2] 
S[table-format=1.2] 
S[table-format=1.2] 
S[table-format=1.2] 
S[table-format=1.2] 
S[table-format=1.2] 
S[table-format=1.2] 
S[table-format=1.2] 
}
\toprule
\multirow{3}{*}{$n$} & \multirow{3}{*}{$p$} & \multirow{3}{*}{$p_{\text{true}}$} & \multicolumn{4}{c}{SpaCoBi} & \multicolumn{4}{c}{Bi-ADMM} & \multicolumn{4}{c}{COBRA} \\
\cmidrule(lr){4-7} \cmidrule(lr){8-11} \cmidrule(lr){12-15}
& & & \multicolumn{2}{c}{Train} & \multicolumn{2}{c}{Val} & \multicolumn{2}{c}{Train} & \multicolumn{2}{c}{Val} & \multicolumn{2}{c}{Train} & \multicolumn{2}{c}{Val} \\
\cmidrule(lr){4-5} \cmidrule(lr){6-7} \cmidrule(lr){8-9} \cmidrule(lr){10-11} \cmidrule(lr){12-13} \cmidrule(lr){14-15}
& & & {Mean} & {SD} & {Mean} & {SD} & {Mean} & {SD} & {Mean} & {SD} & {Mean} & {SD} & {Mean} & {SD} \\
\midrule
\multirow{6}{*}{60} 
& 200 & 40 & 0.82 & 0.13 & 0.84 & 0.12 & 0.77 & 0.12 & 0.80 & 0.10 & 0.15 & 0.18 & 0.15 & 0.18 \\
& 400 & 40 & 0.91 & 0.05 & 0.79 & 0.08 & 0.77 & 0.05 & 0.77 & 0.05 & 0.06 & 0.08 & 0.06 & 0.08 \\
& 600 & 40 & 0.83 & 0.03 & 0.79 & 0.08 & 0.25 & 0.27 & 0.26 & 0.25 & 0.02 & 0.02 & 0.02 & 0.02 \\
\cmidrule(lr){2-15}
& 200 & 60 & 0.83 & 0.14 & 0.86 & 0.13 & 0.79 & 0.15 & 0.82 & 0.14 & 0.50 & 0.25 & 0.52 & 0.26 \\
& 400 & 60 & 0.72 & 0.21 & 0.77 & 0.19 & 0.66 & 0.21 & 0.68 & 0.21 & 0.19 & 0.16 & 0.19 & 0.16 \\
& 600 & 60 & 0.78 & 0.17 & 0.74 & 0.14 & 0.15 & 0.18 & 0.15 & 0.19 & 0.03 & 0.02 & 0.03 & 0.02 \\
\midrule
\multirow{6}{*}{120} 
& 200 & 40 & 0.82 & 0.15 & 0.83 & 0.14 & 0.77 & 0.15 & 0.78 & 0.15 & 0.17 & 0.18 & 0.17 & 0.18 \\
& 400 & 40 & 0.96 & 0.03 & 0.95 & 0.02 & 0.73 & 0.04 & 0.76 & 0.05 & 0.05 & 0.08 & 0.05 & 0.08 \\
& 600 & 40 & 0.75 & 0.03 & 0.75 & 0.03 & 0.20 & 0.21 & 0.20 & 0.21 & 0.03 & 0.02 & 0.03 & 0.02 \\
\cmidrule(lr){2-15}
& 200 & 60 & 0.84 & 0.12 & 0.89 & 0.10 & 0.40 & 0.20 & 0.41 & 0.20 & 0.49 & 0.21 & 0.49 & 0.21 \\
& 400 & 60 & 0.71 & 0.22 & 0.76 & 0.21 & 0.23 & 0.22 & 0.24 & 0.22 & 0.19 & 0.19 & 0.19 & 0.19 \\
& 600 & 60 & 0.94 & 0.01 & 0.96 & 0.02 & 0.12 & 0.15 & 0.12 & 0.15 & 0.09 & 0.12 & 0.09 & 0.12 \\
\midrule
\multirow{6}{*}{240} 
& 200 & 40 & 0.79 & 0.17 & 0.85 & 0.15 & 0.26 & 0.18 & 0.27 & 0.18 & 0.24 & 0.19 & 0.24 & 0.19 \\
& 400 & 40 & 0.72 & 0.03 & 0.72 & 0.03 & 0.25 & 0.24 & 0.25 & 0.24 & 0.06 & 0.07 & 0.06 & 0.07 \\
& 600 & 40 & 0.79 & 0.13 & 0.78 & 0.11 & 0.13 & 0.18 & 0.13 & 0.18 & 0.03 & 0.02 & 0.03 & 0.02 \\
\cmidrule(lr){2-15}
& 200 & 60 & 0.91 & 0.07 & 0.95 & 0.06 & 0.46 & 0.28 & 0.47 & 0.28 & 0.42 & 0.20 & 0.42 & 0.20 \\
& 400 & 60 & 0.84 & 0.15 & 0.85 & 0.15 & 0.30 & 0.22 & 0.30 & 0.22 & 0.16 & 0.19 & 0.19 & 0.17 \\
& 600 & 60 & 0.83 & 0.15 & 0.84 & 0.15 & 0.28 & 0.25 & 0.28 & 0.25 & 0.10 & 0.11 & 0.10 & 0.11 \\
\bottomrule
\end{tabular}
\end{table}

\begin{table}[htbp]
\centering
\caption{Comparison of Feature Selection Performance: False Negative Rate (FNR), False Positive Rate (FPR), and Area Under the Curve (AUC)}
\label{tab:fnr_fpr_auc_comparison_reorganized}
\setlength{\tabcolsep}{3pt} 
\begin{tabular}{
c 
c 
c 
S[table-format=1.2] 
S[table-format=1.2] 
S[table-format=1.2] 
S[table-format=1.2] 
S[table-format=1.2] 
S[table-format=1.2] 
S[table-format=1.2] 
S[table-format=1.2] 
S[table-format=1.2] 
S[table-format=1.2] 
}
\toprule
\multirow{3}{*}{$n$} & \multirow{3}{*}{$p$} & \multirow{3}{*}{$p_{\text{true}}$} & \multicolumn{6}{c}{SpaCoBi} & \multicolumn{4}{c}{Bi-ADMM} \\
\cmidrule(lr){4-9} \cmidrule(lr){10-13}
& & & \multicolumn{2}{c}{FNR} & \multicolumn{2}{c}{FPR} & \multicolumn{2}{c}{AUC} & \multicolumn{2}{c}{FNR} & \multicolumn{2}{c}{FPR} \\
\cmidrule(lr){4-5} \cmidrule(lr){6-7} \cmidrule(lr){8-9} \cmidrule(lr){10-11} \cmidrule(lr){12-13}
& & & {Mean} & {SD} & {Mean} & {SD} & {Mean} & {SD} & {Mean} & {SD} & {Mean} & {SD} \\
\midrule
\multirow{3}{*}{60} 
& 200 & 40 & 0.06 & 0.04 & 0.23 & 0.16 & 0.90 & 0.15 & 0.00 & 0.00 & 1.00 & 0.00 \\
& 400 & 40 & 0.00 & 0.01 & 0.04 & 0.05 & 0.79 & 0.15 & 0.00 & 0.00 & 1.00 & 0.00 \\
& 600 & 60 & 0.02 & 0.03 & 0.13 & 0.16 & 0.89 & 0.16 & 0.00 & 0.00 & 1.00 & 0.00 \\
\midrule
\multirow{2}{*}{120} 
& 200 & 40 & 0.07 & 0.03 & 0.27 & 0.12 & 0.81 & 0.12 & 0.00 & 0.00 & 1.00 & 0.00 \\
& 400 & 60 & 0.03 & 0.04 & 0.16 & 0.22 & 0.78 & 0.13 & 0.00 & 0.00 & 1.00 & 0.00 \\
\midrule
\multirow{2}{*}{240} 
& 200 & 40 & 0.03 & 0.04 & 0.12 & 0.18 & 0.76 & 0.24 & 0.00 & 0.00 & 1.00 & 0.00 \\
& 400 & 40 & 0.01 & 0.02 & 0.12 & 0.18 & 0.88 & 0.18 & 0.00 & 0.00 & 1.00 & 0.00 \\
\bottomrule
\end{tabular}
\end{table}


\subsection{Application to MOB Data}
This study utilized a real-world mouse olfactory bulb (MOB) gene expression dataset, with sample labels determined by the Biotechnology Research Center of the Institute of Advanced Natural Sciences at Beijing Normal University, as a biological genomics application case to evaluate the performance of the Sparse Convex Biclustering algorithm (SpaCoBi) and the Convex Biclustering algorithm (Bi-ADMM) under the \(L_2\)-norm. The original dataset comprises 305 observed samples and 1,250 gene features. As illustrated in Figure \ref{Fig.main2}, the heatmap of the raw data reveals two salient characteristics: first, a distinct vertical stripe pattern, suggesting that certain subsets of samples may significantly influence clustering outcomes; second, extensive regions with near-zero expression values, indicating a high degree of sparsity and the presence of numerous uninformative features. These high-dimensional and highly sparse characteristics motivated the application of the SpaCoBi algorithm, which exploits sparsity to uncover underlying clustering structures and identify critical genetic markers.

\begin{figure}
		\centering 
		\includegraphics[width=0.7\textwidth]{ 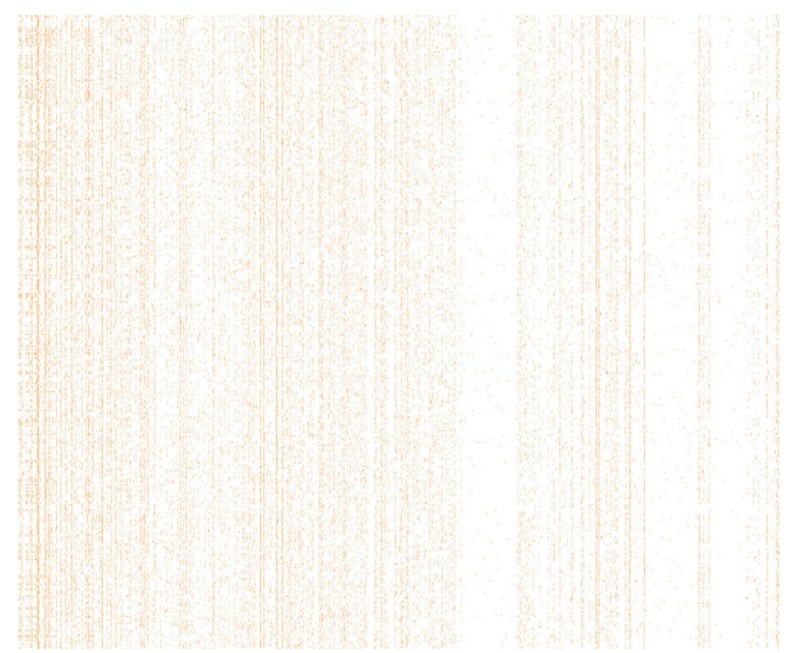} 
		\caption{The heat map of the original data.} 
		\label{Fig.main2} 
\end{figure}


To rigorously evaluate the practical utility of the SpaCoBi and Bi-ADMM (\(L_2\)-norm) algorithms in genomic data analysis, we leveraged the known biological classifications of the 305 samples. By comparing the clustering outcomes from both algorithms against the ground-truth biological classes, we quantitatively assessed their accuracy. Both algorithms were executed with identical penalty parameter settings to ensure a fair comparison. Figures \ref{Fig.main3} and \ref{Fig.main4} present the heatmaps of the clustering results obtained from SpaCoBi and Bi-ADMM, respectively. The comparative analysis indicates that the heatmap generated by SpaCoBi demonstrates clearer delineation of clusters and effectively suppresses uninformative features, thereby highlighting the advantages of its sparsity-inducing penalty. Furthermore, the clustering structure derived from SpaCoBi closely reflects the known three-class organization of the samples, while Bi-ADMM struggles to distinguish between these classes. This observation is quantitatively supported by the Adjusted Rand Index (ARI), which reaches 1.0 for SpaCoBi, in stark contrast to a mere 0.12 for Bi-ADMM.

\begin{figure}
		\centering 
		\includegraphics[width=0.7\textwidth]{ 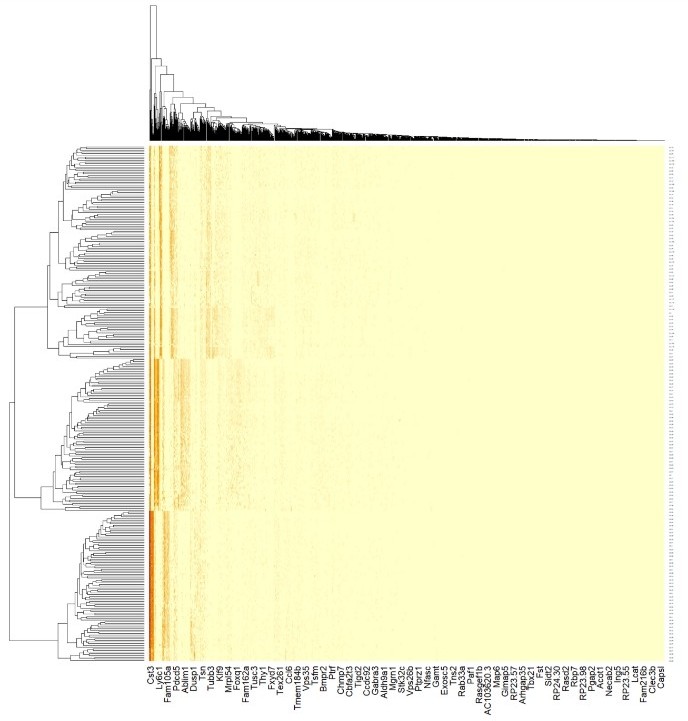} 
		\caption{The heat map of $\hat{A}$ estimated by the SpaCoBi algorithm} 
		\label{Fig.main3} 
\end{figure}

\begin{figure}
		\centering 
		\includegraphics[width=0.7\textwidth]{ 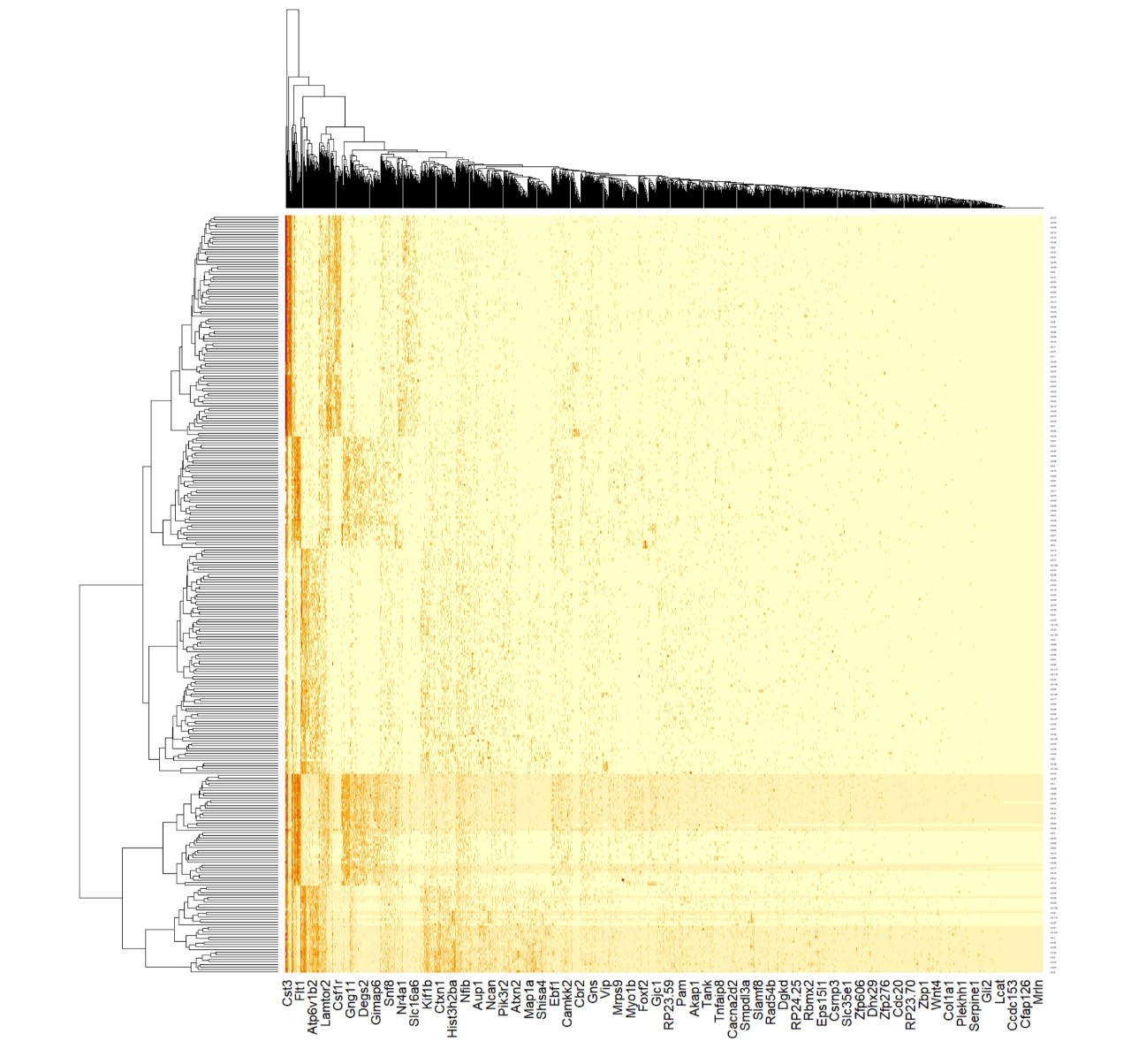} 
		\caption{The heat map of $\hat{A}$ estimated by the Bi-ADMM algorithm} 
		\label{Fig.main4} 
\end{figure}

An ARI of 1.0 for the SpaCoBi algorithm provides compelling evidence that it accurately captures the intrinsic clustering structure within high-dimensional biological gene expression data, with classification results perfectly aligning with the true biological annotations. This significant improvement is attributed to SpaCoBi’s capacity to identify and mitigate the effects of irrelevant or noisy features through its inherent sparsity mechanism, thereby enhancing classification accuracy. According to the feature selection outputs from SpaCoBi, the key gene features contributing significantly to the clustering include: ``Pbxip1", ``Pdlim2", ``Cdc34", ``Kdm7a", ``Ptprz1", ``Kctd13", ``Higd1b", ``Bcas1", ``Gpcpd1", ``Man2b2", ``Inpp4a", ``Mef2c", ``Ftsj3", ``Flii", ``Osr1", ``Slc39a1", ``Armc6", ``label", ``Nell2", ``RP23.96", ``Car4", ``Epb41l5", and ``Isg15". This identified subset of informative genes provides valuable insights and targeted avenues for future experimental validation and mechanistic studies in molecular biology.

\section{Conclusion}
In this manuscript, we proposed the Sparse Convex Biclustering algorithm as a robust method for analyzing high-dimensional data. Through comprehensive simulations and a real-world application using a mouse olfactory bulb gene expression dataset, we demonstrated SpaCoBi's substantial advantages over existing convex biclustering algorithms.

The results based on the MOB data indicated that SpaCoBi significantly outperformed Bi-ADMM in terms of clustering accuracy, as evidenced by an Adjusted Rand Index (ARI) of 1.0, which reflects its ability to accurately capture the intrinsic clustering structure of the data. This efficiency stems from SpaCoBi’s inherent sparsity mechanism, which effectively identifies and removes the influence of irrelevant or noisy features, thus enhancing classification precision. Furthermore, the feature selection capabilities of SpaCoBi highlighted key informative genes such as "Pbxip1", "Pdlim2", "Cdc34", and others, providing valuable insights into the biological processes underlying the data. This identified subset of genes offers targeted directions for future experimental validation and mechanistic studies in molecular biology.
 Overall, this research underscores the importance of selecting informative features in high-dimensional settings and illustrates how the SpaCoBi algorithm can serve as a powerful tool for genomic data analysis.

\bibliographystyle{apalike}
\bibliography{refs_clustering.bib}

\section{Appendix}

{\bf Details for deriving the update of $\mbfA$ in Section 2.2 Step 1:}

Note that $A_{l_1\cdot}-A_{l_2\cdot}= \mbfA\trans(\mbfe_{l_1}-\mbfe_{l_2})$, $\mbfa_{k_1}-\mbfa_{k_2}= \mbfA(\mbfe_{k_1}^*-\mbfe_{k_2}^*)$ and $\mbfa_j = \mbfA \mbfe_j^*$, where $\mbfe_{l_1}$ is a $n$-dimensional vector with its $l_1$-th element as 1 and otherwise as 0,  and $\mbfe_{k_1}^*$ is a $p$-dimensional vector with its $k_1$-th element as 1 and otherwise as 0. By vectorizing matrices $\boldsymbol{a} = \textrm{vec}(\mbfA)$ and applying the identity
\begin{eqnarray*}
\textrm{vec}(\mbfR \mbfS \mbfT)=[\mbfT\trans \otimes \mbfR ] \textrm{vec}(\mbfS),
\end{eqnarray*}
it follows
\begin{eqnarray*}
f(\boldsymbol{a})=\frac{1}{2} \| \mbfx- \boldsymbol{a} \|_2^2 + \frac{\nu_1}{2} \sum_{l \in \calE_l} \| \mbfB_l \mbfP \boldsymbol{a}  - \wt{\mbfv}_l  \|_2^2  +\frac{\nu_2}{2} \sum_{k \in \calE_2} \| \mbfC_k  \boldsymbol{a}  -\wt{\mbfz}_k \|_2^2 + \frac{\nu_3}{2} \sum_{j=1}^p ( \mbfD_j  \boldsymbol{a}  - \wt{\mbfg}_j )^2,
\end{eqnarray*}
where
\begin{eqnarray*}
\mbfB_l&=&(\mbfe_{l_1}-\mbfe_{l2})\trans \otimes \mbfI_p,\ 
\mbfC_k=(\mbfe_{k_1}^*-\mbfe_{k2}^*)\trans \otimes \mbfI_n \\
\mbfD_i&=& (\mbfe_j^*)\trans \otimes \mbfI_n,\ 
\textrm{vec}(\mbfA\trans)=\mbfP \textrm{vec}(\mbfA).
\end{eqnarray*}
With a little abuse of notations, note that $\mbfP=(P_{kl}),1\leq k,l \leq np$ here is a unique permutation matrix such that $P_{kl}=1$ if $k=(i-1)n+j$ and $l=(j-1)p+i, 1\leq i \leq p, 1\leq j \leq n$, and 0 otherwise. It is easy to see $\mbfP\trans=\mbfP^{-1}$. Let $\varepsilon_1=|\calE_1|,\varepsilon_2=|\calE_2|$, and 
\begin{eqnarray*}
&&\mbfB\trans=\left(\mbfB\trans_1,\ldots,\mbfB\trans_{\varepsilon_1} \right), \quad \wt{\mbfv}\trans= \left(\wt{\mbfv}_1\trans,\ldots,\wt{\mbfv}_{\varepsilon_1}\trans \right) \\
&&\mbfC\trans=\left(\mbfC\trans_1,\ldots,\mbfC\trans_{\varepsilon_2} \right), \quad \wt{\mbfz}\trans=
\left( \wt{\mbfz}_1\trans,\ldots,\wt{\mbfz}_{\varepsilon_2}\trans \right) \\
&&\mbfD\trans=\left(\mbfD\trans_1,\ldots,\mbfD\trans_n \right), \quad \wt{\mbfg}\trans=
\left( \wt{\mbfg}_1\trans,\ldots,\wt{\mbfg}_p\trans \right).
\end{eqnarray*}

Then we have
\begin{eqnarray*}
f(\boldsymbol{a})=\frac{1}{2} \| \mbfx- \boldsymbol{a} \|_2^2 + \frac{\nu_1}{2} \| \mbfB \mbfP \boldsymbol{a} - \wt{\mbfv}  \|_2^2  +\frac{\nu_2}{2}  \| \mbfC \boldsymbol{a} -\wt{\mbfz} \|_2^2 + \frac{\nu_3}{2}  \| \mbfD \boldsymbol{a} -\wt{\mbfg} \|_2^2.
\end{eqnarray*}

The stationary equation can be obtained by
\begin{eqnarray*}
(\mbfI_{np} + \nu_1 \mbfP\trans \mbfB\trans \mbfB \mbfP + \nu_2 \mbfC\trans \mbfC + \nu_3 \mbfD\trans \mbfD) \boldsymbol{a} = \mbfx +\nu_1 \mbfP\trans \mbfB\trans \wt{\mbfv} + \nu_2 \mbfC\trans \wt{\mbfz} + \nu_3 \mbfD\trans\wt{\mbfg}.
\end{eqnarray*}

This is a system of \( np \) linear equations. We can attempt to simplify its form by applying properties of the Kronecker product, such as
 $(\mbfS\otimes \mbfT)\trans=\mbfS\trans \otimes \mbfT\trans$ and $(\mbfQ \otimes \mbfR)(\mbfS\otimes \mbfT)=(\mbfQ\mbfS) \otimes (\mbfR \mbfT)$.Then, it follows
\begin{eqnarray*}
\nu_2 \mbfC\trans \mbfC &=& \nu_2 \sum_{k \in \calE_2} \left[ \left((\mbfe_{k_1}^*-\mbfe_{k_2}^*) (\mbfe_{k_1}^*-\mbfe_{k_2}^*)\trans \right) \right] \otimes \mbfI_n \\
&=& \left[\sum_{k \in \calE_2}\nu_2 \left((\mbfe_{k_1}^*-\mbfe_{k_2}^*) (\mbfe_{k_1}^*-\mbfe_{k_2}^*)\trans \right) \otimes \right]  \mbfI_n \\
\nu_2\mbfC\trans \wt{\mbfz} &=& \nu_2 \sum_{k \in \calE_2} [(\mbfe_{k_1}^*-\mbfe_{k_2}^*) \otimes \mbfI_n ] \wt{\mbfz}_k = \nu_2\sum_{k \in \calE_2} \left((\mbfe_{k_1}^*-\mbfe_{k_2}^*)  \otimes  \mbfI_n \right) \wt{\mbfz}_k \\
\nu_3 \mbfD\trans \mbfD &=& \nu_3 \sum_{j=1}^p \mbfD_i\trans \mbfD_i = \left[ \nu_3 \sum_{j=1}^p \mbfe_j^* (\mbfe_j^*)\trans \right] \otimes \mbfI_n \\
\nu_3\mbfD\trans \wt{\mbfg} &=& \nu_3 \sum_{j=1}^p \mbfD_j \trans \wt{\mbfg}_j = \nu_3 \sum_{j=1}^p \left( \mbfe_j^* \otimes \mbfI_n \right)\wt{\mbfg}_j.
\end{eqnarray*}
Here, we apply the properties of \( \mbfP \) shown in Proposition 1 and it can be obtained
\begin{eqnarray*}
\mbfI_{np} + \nu_1 \mbfP\trans \mbfB\trans \mbfB \mbfP &=& \mbfI_p \otimes \left[ \mbfI_n + \nu_1 \sum_{l \in \calE_1} (\mbfe_{l_1}-\mbfe_{l_2})(\mbfe_{l_1}-\mbfe_{l_2})\trans \right] \\
\nu_1 \mbfP\trans \mbfB\trans \wt{\mbfv}&=&  \sum_{l \in \calE_1}\left[ \left(\mbfI_p \otimes \nu_1(\mbfe_{l_1}-\mbfe_{l_2}) \right) \wt{\mbfv}_l \right].
\end{eqnarray*}

Therefore, the system of equations is equivalent to
\begin{eqnarray*} 
&& \ \ (\mbfI_p \otimes \mbfM)\textrm{vec}(\mbfA) + (\mbfN\otimes \mbfI_n )\textrm{vec}(\mbfA) = \textrm{vec}(\mbfH)  \nonumber\\
&\Leftrightarrow& \ \ \mbfM\mbfA + \mbfA\mbfN = \mbfH. \ \ \ \ \hfill{} \blacksquare
\end{eqnarray*}

\begin{proposition}
For the permutation matrix $\mbfP$ defined above, we can prove for any $k \in \calE_2$ and $p$-dimensional vector $\mbfd$,
\begin{enumerate}
\item[(1).] $\left[ \mbfd\trans \otimes \mbfI_p \right]\mbfP =\mbfI_p \otimes \mbfd\trans$;

\item[(2).] $\mbfP\trans \left[ \left(\mbfd\mbfd\trans \right) \otimes \mbfI_p \right]\mbfP =\mbfI_p \otimes \left(\mbfd\mbfd\trans \right)$.
\end{enumerate}
\end{proposition}

The proof of Proposition 1 is shown below:

\begin{enumerate}
\item[(1).] Note that
$\bfP=(P_{kl}),1\leq k,l\leq np$ here is a unique permutation matrix
such that $P_{kl}=1$ if $k=(i-1)p+j$ and $l=(j-1)n+i,1\leq i\leq n,1\leq j\leq p$,
and 0 otherwise. By the definition of $\bfP$, it is clear that multiplying
a matrix by $\bfP$ on the right moves its $k$-th column to the $l$-th
column when $P_{kl}=1$.

Consider the $i$-th element $d_{i}$ of $\bfd$, then in $\bfd\trans\otimes\mbfI_{p}$,
its entries at $(j,(i-1)p+j)$ equal $d_{i}$, $j=1,\ldots,p$. Thus,
in $(\bfd\trans\otimes\mbfI_{n})\bfP$, the entry at $(j,(j-1)n+i)$
equals to $d_{i}$. In $\mbfI_{p}\otimes\bfd\trans$, it is easy to
see the entry at $(j,(j-1)n+i)$ equal $d_{i},i=1,\ldots,n,j=1,\ldots,p$.

\item[(2).] \begin{eqnarray*}
\mbfP\trans \left[ \left(\mbfd\mbfd\trans \right) \otimes \mbfI_p \right]\mbfP &=& \mbfP\trans \left[ (\mbfd \otimes \mbfI_p ) (\mbfd\trans \otimes \mbfI_p) \right]\mbfP \\
&=& \left[(\mbfd\trans \otimes \mbfI_p ) \mbfP  \right]\trans \left[ (\mbfd\trans \otimes \mbfI_p ) \mbfP \right]\\
&=& (\mbfI_p \otimes \mbfd\trans)\trans (\mbfI_p \otimes \mbfd\trans) \\
&=& \mbfI_p \otimes \left(\mbfd\mbfd\trans \right). \hfill{} \ \ \ \ \blacksquare
\end{eqnarray*}

\end{enumerate}


\end{document}